\documentclass[11pt,a4paper]{article}
\usepackage[hyperref]{AACL-IJCNLP2020}
\usepackage{times}
\usepackage{latexsym}
\usepackage{multirow}
\usepackage{booktabs}
\usepackage{colortbl}
\usepackage{subcaption}

\usepackage{todonotes}

\newcommand{\squishlist}{
 \begin{list}{$\bullet$}
  { \setlength{\itemsep}{0pt}
     \setlength{\parsep}{1pt}
     \setlength{\topsep}{1pt}
     \setlength{\partopsep}{0pt}
     \setlength{\leftmargin}{2em}
     \setlength{\labelwidth}{0.5em}
     \setlength{\labelsep}{0.5em} } }
 \newcommand{\squishend}{\end{list}}
\newcommand{\eat}[1]{}

\usepackage{microtype}

\aclfinalcopy 

\usepackage{xspace}

\title{A Survey of the State of Explainable AI for Natural Language Processing}

\author{Marina Danilevsky, Kun Qian, Ranit Aharonov, Yannis Katsis, Ban Kawas, Prithviraj Sen  \\
  IBM Research -- Almadem\\
  \texttt{mdanile@us.ibm.com}, \texttt{\{qian.kun,Ranit.Aharonov2\}@ibm.com}\\
  \texttt{yannis.katsis@ibm.com}, \texttt{\{bkawas,senp\}@us.ibm.com}\\}

\date{}

\begin{document}
\maketitle

\linepenalty=1000

\begin{abstract}
Recent years have seen important advances in the quality of state-of-the-art models, but this has come at the expense of models becoming less interpretable.
This survey presents an overview of the current state of Explainable AI (XAI), considered within the domain of Natural Language Processing (NLP). 
We discuss the main categorization of explanations, as well as the various ways explanations can be arrived at and visualized.
We detail the operations and explainability techniques currently available for generating explanations for NLP model predictions, to serve as a resource for model developers in the community. 
Finally, we point out the current gaps and encourage directions for future work in this important research area.
\end{abstract}

\section{Introduction}
\label{sec:intro}

Traditionally, Natural Language Processing (NLP) systems have been mostly based on techniques that are inherently explainable. Examples of such approaches, often referred to as \emph{white box} techniques, include rules, decision trees, hidden Markov models, logistic regressions, and others.
Recent years, though, have brought the advent and popularity of \emph{black box} techniques, such as deep learning models and the use of language embeddings as features. While these methods in many cases substantially advance model quality, they come at the expense of models becoming less interpretable. This obfuscation of the process by which a model arrives at its results can be problematic, as it may erode trust in the many AI systems humans interact with daily (e.g., chatbots, recommendation systems, information retrieval algorithms, and many others).
In the broader AI community, this growing understanding of the importance of explainability has created an emerging field called Explainable AI (XAI). 
However, just as tasks in different fields are more amenable to particular approaches, explainability must also be considered within the context of each discipline. 
We therefore focus this survey on 
XAI works in the domain of NLP, as represented in the main NLP conferences in the last seven years. This is, to the best of our knowledge, the first XAI survey focusing on the NLP domain.

As will become clear in this survey, explainability is in itself a term that requires an explanation. While explainability may generally serve many purposes \cite[see, e.g.,][] {lertvittayakumjorn-toni-2019-human}, 
our focus is on explainability from the perspective of an end user whose goal is to understand how a model arrives at its result, also referred to as the \textit{outcome explanation problem} \cite{guidotti-2019-acm-exp-bbm}. 
In this regard, explanations can help users of NLP-based AI systems build trust in these systems' predictions. Additionally, understanding the model's operation may also allow users to provide useful feedback, which in turn can help developers improve model quality \cite{adadi-2018-ieee-peeking}.

Explanations of model predictions have previously been categorized in a fairly simple way that differentiates between (1) whether the explanation is for each prediction individually or the model's prediction process as a whole, and (2) determining whether generating the explanation requires post-processing or not (see Section~\ref{sec:what-and-when-explain}).
However, although rarely studied, there are many additional characterizations of explanations, the most important being the techniques used to either generate or visualize explanations. In this survey, we analyze the NLP literature with respect to both these dimensions and identify the most commonly used \emph{explainability and visualization techniques}, in addition to \emph{operations} used to generate explanations (Sections~\ref{sec:exp-techniques}-Section~\ref{sec:viz-techniques}).
We briefly describe each technique and point to representative papers adopting it.
Finally, we discuss the common \emph{evaluation techniques} used to measure the quality of explanations (Section~\ref{sec:how-well}), and conclude with a discussion of gaps and challenges in developing successful explainability approaches in the NLP domain (Section~\ref{sec:conclusion}). 
\noindent {\bf Related Surveys}: Earlier surveys on XAI include \citet{adadi-2018-ieee-peeking} and \citet{guidotti-2019-acm-exp-bbm}. While \citeauthor{adadi-2018-ieee-peeking} provide a comprehensive review of basic terminology and fundamental concepts relevant to XAI in general, our goal is to survey more recent works in NLP in an effort to understand how these achieve XAI and how well they achieve it. \citeauthor{guidotti-2019-acm-exp-bbm} adopt a four dimensional classification scheme to rate various approaches. Crucially, they differentiate between the ``explanator" and the black-box model it explains. This makes most sense when a surrogate model is used to explain a black-box model. As we shall subsequently see, such a distinction applies less well to the majority of NLP works published in the past few years where the same neural network (NN) can be used not only to make predictions but also to derive explanations. In a series of tutorials, \citet{lecue-2020-xai-tutorial} discuss fairness and trust in machine learning (ML) that are clearly related to XAI but not the focus of this survey. Finally, we adapt some nomenclature from \citet{Arya2019OneED} which presents a software toolkit that can help users lend explainability to their models and ML pipelines.

Our goal for this survey is to: (1) provide the reader with a better understanding of the state of XAI in NLP, (2) point developers interested in building explainable NLP models to currently available techniques, and (3) bring to the attention of the research community the gaps that exist; mainly a lack of formal definitions and evaluation for explainability. We have also built an interactive website providing interested readers with all relevant aspects for every paper covered in this survey. \footnote{\url{https://xainlp2020.github.io/xainlp/} (we plan to maintain this website as a contribution to the community.)}



\section{Methodology}
\label{sec:methodology}

We identified relevant papers (see Appendix \ref{sec:appendix-methodology}) and classified them based on the aspects defined in Sections~\ref{sec:what-and-when-explain} and \ref{sec:how-explain}. To ensure a consistent classification, each paper was individually analyzed by at least two reviewers, consulting additional reviewers in the case of disagreement. For simplicity of presentation, we label each paper with its main applicable category for each aspect, though some papers may span multiple categories (usually with varying degrees of emphasis.)
All relevant aspects for every paper covered in this survey can be found at the aforementioned website; to enable readers of this survey to discover interesting explainability techniques and ideas, even if they have not been fully developed in the respective publications.

\section{Categorization of Explanations}
\label{sec:what-and-when-explain}

Explanations are often categorized along two main aspects \cite{guidotti-2019-acm-exp-bbm,adadi-2018-ieee-peeking}. The first distinguishes whether the explanation is for an individual prediction (\emph{local}) or the model's prediction process as a whole (\emph{global}). The second differentiates between the explanation emerging directly from the prediction process (\emph{self-explaining}) versus requiring post-processing (\emph{post-hoc}). We next describe both of these aspects in detail, and provide a summary of the four categories they induce in Table~\ref{tab:local-global-self-post}.

\subsection{Local vs Global} 
\label{sec:what-explain}

A \textit{local} explanation provides information or justification for the model's prediction on a specific input; 46 of the 50 papers fall into this category. 




A \textit{global} explanation provides similar justification by revealing how the model's predictive process works, independently of any particular input. This category holds the remaining 4 papers covered by this survey. 
This low number is not surprising given the focus of this survey being on explanations that justify predictions, as opposed to explanations that help understand a model's behavior in general (which lie outside the scope of this survey).







\subsection{Self-Explaining vs Post-Hoc} 
\label{sec:when-explain}  


Regardless of whether the explanation is local or global, explanations differ on whether they 
 arise as part of the prediction process, or whether their generation requires post-processing following the model making a prediction.
A \textit{self-explaining} approach, which may also be referred to as directly interpretable \cite{Arya2019OneED}, generates the explanation at the same time as the prediction, using information emitted by the model as a result of the process of making that prediction. 
Decision trees and rule-based models are examples of global self-explaining models, while feature saliency approaches such as attention are examples of local self-explaining models.


In contrast, a \textit{post-hoc} approach requires that an additional operation is performed after the predictions are made. 
LIME \cite{ribeiro:kdd16} is an example of producing a local explanation using a surrogate model applied following the predictor's operation. 
A paper might also be considered to span both categories -- for example, \cite{sydorova-etal-2019-interpretable} actually presents both self-explaining and post-hoc explanation techniques.




\begin{table}[!h]
\centering
\small
\begin{tabular}{p{1.6cm}p{5cm}}
\toprule
\textbf{Local \mbox{Post-Hoc}} & Explain a single prediction by performing additional operations (\textit{after} the model has emitted a prediction) \\
\arrayrulecolor{gray}\midrule
\textbf{Local Self- Explaining} & Explain a single prediction using the model itself (calculated from information made available from the model \textit{as part of} making the prediction) \\
\arrayrulecolor{gray}\midrule
\textbf{Global \mbox{Post-Hoc}} & Perform additional operations to explain the entire model's predictive reasoning \\
\arrayrulecolor{gray}\midrule
\textbf{Global Self- Explaining} & Use the predictive model itself to explain the entire model's predictive reasoning (\textit{a.k.a.} directly interpretable model) \\
\bottomrule
\end{tabular}
\caption{ Overview of the high-level categories of explanations (Section~\ref{sec:what-and-when-explain}).}
\label{tab:local-global-self-post}
\end{table}





\section{Aspects of Explanations}
\label{sec:how-explain}


\begin{table*}[!ht]
\centering
\small
\begin{tabular}{p{1.4cm}p{2cm}p{4.5cm}p{2cm}p{0.25cm}p{3.2cm}}
\toprule
\textbf{Category} & \textbf{Explainability} & \textbf{Operations to} & \textbf{Visualization} &  & \textbf{Representative} \\
\textbf{(\#)}  & \textbf{Technique} & \textbf{Enable Explainability} & \textbf{ Technique} & \textbf{\#} & \textbf{Paper(s)} \\
  \arrayrulecolor{black} \midrule
\textbf{Local \mbox{Post-Hoc}} \textit{(11)} & feature \mbox{importance} & first derivative saliency, example driven & saliency & 5 & \citep{wallace-etal-2018-interpreting,ijcai2017-371} \\
\arrayrulecolor{gray}\cmidrule{2-6}
 & surrogate model & first derivative saliency, layer-wise relevance propagation, input perturbation & saliency & 4 & \citep{alvarez-melis-jaakkola-2017-causal,poerner-etal-2018-evaluating,ribeiro:kdd16} \\
 \arrayrulecolor{gray}\cmidrule{2-6}
 & example driven & layer-wise relevance propagation, explainability-aware architecture & raw examples & 2 & \citep{croce-etal-2018-explaining,jiang-etal-2019-explore} \\
 \arrayrulecolor{black} \midrule
\textbf{Local \mbox{Self-Exp}} \textit{(35)} & feature \mbox{importance} & attention, first derivative saliency, LSTM gating signals, explainability-aware architecture & saliency & 22 & \cite{mullenbach-etal-2018-explainable, ghaeini-etal-2018-interpreting, xie-etal-2017-interpretable, aubakirova:emnlp16} \\
\arrayrulecolor{gray}\cmidrule{2-6}
 & induction & explainability-aware architecture, rule induction  & raw declarative representation & 6 & \citep{ling-etal-2017-program,dong-etal-2019-editnts,pezeshkpour-etal-2019-investigating} \\
 \arrayrulecolor{gray}\cmidrule{2-6}
  & provenance & template-based & natural \mbox{language}, other & 3 & \citep{abujabal2017quint} \\
 \arrayrulecolor{gray}\cmidrule{2-6}
 & surrogate model & attention, input perturbation, explainability-aware architecture & natural \mbox{language} & 3 & \citep{rajani-etal-2019-explain,sydorova-etal-2019-interpretable}\\
 \arrayrulecolor{gray}\cmidrule{2-6}
 & example driven & layer-wise relevance propagation & raw examples & 1 & \citep{croce-etal-2019-auditing} \\
 \arrayrulecolor{black} \midrule
\textbf{Global} \mbox{\textbf{Post-Hoc}} \textit{(3)} & feature \mbox{importance} & class activation mapping, attention, gradient reversal & saliency & 2 & \citep{pryzant-etal-2018-interpretable,pryzant-etal-2018-deconfounded} \\
\arrayrulecolor{gray}\cmidrule{2-6}
 & surrogate   model & taxonomy induction & raw declarative representation & 1 & \citep{liu-kdd-18}\\
 \arrayrulecolor{black} \midrule
\textbf{Global \mbox{Self-Exp}} \textit{(1)} & induction & reinforcement learning & raw declarative representation & 1 & \citep{prollochs-etal-2019-learning} \\
\bottomrule
\end{tabular}
\caption{
Overview of common combinations of explanation aspects: columns 2, 3, and 4 capture explainability techniques, operations, and visualization techniques, respectively (see Sections \ref{sec:exp-techniques}, \ref{sec:operations}, and \ref{sec:viz-techniques} for details). These are grouped by the high-level categories detailed in Section \ref{sec:what-and-when-explain}, as shown in the first column. The last two columns show the number of papers in this survey that fall within each subgroup, and a list of representative references.}
\label{tab:aspects-papers}
\end{table*}

While the previous categorization serves as a convenient high-level classification of explanations, it does not cover other important characteristics.
We now introduce two additional aspects of explanations: 
(1) techniques for deriving the explanation and (2) presentation to the end user. We discuss the most commonly used explainability techniques, along with basic operations that enable explainability, as well as the visualization techniques commonly used to present the output of associated explainability techniques. 
We identify the most common combinations of explainability techniques, operations, and visualization techniques for each of the four high-level categories of explanations presented above, and summarize them, together with representative papers, in
Table \ref{tab:aspects-papers}.

Although explainability techniques and visualizations are often intermixed, there are fundamental differences between them that motivated us to treat them separately. Concretely, explanation derivation - typically done by AI scientists and engineers - focuses on mathematically motivated justifications of models' output, leveraging various explainability techniques to produce ``raw explanations" (such as attention scores).  On the other hand, explanation presentation - ideally done by UX engineers - focuses on how these ``raw explanations" are best presented to the end users using suitable visualization techniques (such as saliency heatmaps). 


\subsection{Explainability Techniques}
\label{sec:exp-techniques}


In the papers surveyed, we identified five major explainability techniques that differ in the mechanisms they adopt to generate the raw mathematical justifications that lead to the final explanation presented to the end users. 

\textit{Feature importance.} The main idea is to derive explanation by investigating the importance scores of different features used 
to output the final prediction.  
Such approaches can be built on different types of features,
such as manual features obtained from feature engineering \cite[e.g.,][]{voskarides-etal-2015-learning}, 
lexical features including word/tokens and n-gram \cite[e.g.,][]{godin-etal-2018-explaining, mullenbach-etal-2018-explainable},
or latent features learned by NNs \cite[e.g.,][]{xie-etal-2017-interpretable}.
Attention mechanism \cite{bahdanau:iclr15} and first-derivative saliency \cite{li2015visualizing} are two widely used operations to enable feature importance-based explanations. Text-based features are inherently more interpretable by humans than general features, which may explain the widespread use of attention-based approaches in the NLP domain. 

\textit{Surrogate model.} Model predictions are explained by learning a second, usually more explainable model, 
as a proxy. One well-known 
example is LIME \cite{ribeiro:kdd16}, which learns surrogate models using an operation called input perturbation. Surrogate model-based approaches are model-agnostic and can be used to achieve either local \cite[e.g.,][]{alvarez-melis-jaakkola-2017-causal} or global \cite[e.g.,][]{liu-kdd-18} explanations. However, the learned surrogate models and the original models may have completely different mechanisms to make predictions, leading to concerns about the fidelity of surrogate model-based approaches.

\textit{Example-driven.} Such approaches explain the prediction of an input instance by identifying and presenting other instances, usually from available labeled data, that are semantically similar to the input instance. 
They are similar in spirit to nearest neighbor-based approaches \cite{dudani1976distance}, and have been applied to different NLP tasks such as text classification \citep{croce-etal-2019-auditing} and question answering \cite{abujabal2017quint}.

\textit{Provenance-based.} Explanations are provided by illustrating some or all of the prediction derivation process, which is an intuitive and effective explainability technique when the final prediction is the result of a series of reasoning steps. We observe several question answering papers adopt such approaches \citep{abujabal2017quint, zhou2018interpretable, amini-etal-2019-mathqa}.

\textit{Declarative induction.} Human-readable representations, such as rules \cite{prollochs-etal-2019-learning}, trees \cite{voskarides-etal-2015-learning}, and programs \cite{ling-etal-2017-program} are induced as explanations.


As shown in Table \ref{tab:aspects-papers}, feature importance-based and surrogate model-based approaches have been in frequent use (accounting for 29 and 8, respectively, of the 50 papers reviewed). This should not come as a surprise, as features serve as building blocks for machine learning models (explaining the proliferation of feature importance-based approaches) and most recent NLP papers employ NN-based models, which are generally black-box models (explaining the popularity of surrogate model-based approaches). Finally note that a complex NLP approach consisting of different components may employ more than one of these explainability techniques. A representative example is the QA system QUINT \cite{abujabal2017quint}, which displays the query template that best matches the user input query (example-driven) as well as the instantiated knowledge-base entities (provenance).

\subsection{Operations to Enable Explainability}
\label{sec:operations}

We now present the most common set of operations encountered in our literature review that are used to enable explainability, in conjunction with relevant work employing each one.



\textit{First-derivative saliency.} Gradient-based explanations estimate the contribution of input $i$ towards output $o$ by computing the partial derivative of $o$ with respect to $i$. This is closely related to older concepts such as sensitivity \citep{saltelli:book08}. First-derivative saliency is particularly convenient for NN-based models because these can be computed for any layer using a single call to auto-differentiation, which most deep learning engines provide out-of-the-box. Recent work has also proposed improvements to first-derivative saliency \citep{sundararajan:icml18}. 
As suggested by its name and definition, first-derivative saliency can be used to enable feature importance explainability, especially on word/token-level features \cite{aubakirova:emnlp16, karlekar:naacl18}.

\textit{Layer-wise relevance propagation.} This is another way to attribute relevance to features computed in any intermediate layer of an NN. Definitions are available for most common NN layers including fully connected layers, convolution layers and recurrent layers. 
Layer-wise relevance propagation has been used to, for example, enable feature importance explainability \cite{poerner-etal-2018-evaluating} and example-driven explainability \cite{croce-etal-2018-explaining}.

\textit{Input perturbations.} Pioneered by LIME \citep{ribeiro:kdd16}, input perturbations can explain the output for input $\mathbf{x}$ by generating random perturbations of $\mathbf{x}$ and training an explainable model (usually a linear model). 
They are mainly used to enable surrogate models \cite[e.g.,][]{ribeiro:kdd16, alvarez-melis-jaakkola-2017-causal}.

\textit{Attention} \citep{bahdanau:iclr15,vaswani:nips17}. Less an operation and more of a strategy to enable the NN to explain predictions, attention layers can be added to most NN architectures and, because they appeal to human intuition, can help indicate where the NN model is ``focusing". While previous work has widely used attention layers  \citep{luo:ijcai18,xie-etal-2017-interpretable,mullenbach-etal-2018-explainable} to enable feature importance explainability, the jury is still out as to how much explainability attention provides \citep{jain-wallace-2019-attention,serrano-smith-2019-attention,wiegreffe-pinter-2019-attention}.

\textit{LSTM gating signals.} Given the sequential nature of language, recurrent layers, in particular LSTMs \citep{hochreiter:neurcomp97}, are commonplace. While it is common to mine the outputs of LSTM cells to explain outputs, there may also be information present in the outputs of the gates produced within the cells.
It is possible to utilize (and even combine) other operations presented here to interpret gating signals to aid feature importance explainability \citep{ghaeini-etal-2018-interpreting}.

\textit{Explainability-aware architecture design.} One way to exploit the flexibility of deep learning is to devise an NN architecture that mimics the process humans employ to arrive at a solution. This makes the learned model (partially) interpretable since the architecture contains human-recognizable components. 
Implementing such a model architecture can be used to enable the induction of human-readable programs for solving math problems \cite{amini-etal-2019-mathqa,ling-etal-2017-program} or sentence simplification problems \cite{dong-etal-2019-editnts}. This design may also be applied to surrogate models that generate explanations for predictions \cite{rajani-etal-2019-explain,liu-etal-2019-towards-explainable}.


Previous works have also attempted to compare these operations in terms of efficacy with respect to specific NLP tasks \citep{poerner-etal-2018-evaluating}. 
Operations outside of this list exist and are popular for particular categories of explanations. Table \ref{tab:aspects-papers} mentions some of these. For instance, \citet{prollochs-etal-2019-learning} use reinforcement learning to learn simple negation rules, \citet{liu-kdd-18} learns a taxonomy post-hoc to better interpret network embeddings, and \citet{pryzant-etal-2018-deconfounded} uses gradient reversal \citep{ganin:jmlr16} to deconfound lexicons.

\subsection{Visualization Techniques}
\label{sec:viz-techniques}

An explanation may be presented in different ways to the end user, and making the appropriate choice is crucial for the overall success of an XAI approach. For example, the widely used attention mechanism, which learns the importance scores of a set of features, can be visualized as raw attention scores or as a saliency heatmap (see Figure \ref{fig:saliency-heatmap}). Although the former is acceptable, the latter is more user-friendly and has become the standard way to visualize attention-based approaches. We now present the major visualization techniques identified in our literature review.

{
\begin{figure*}[ht!]
\centering
    \begin{subfigure}[t]{0.3\textwidth}
        \centering
        \includegraphics[width=\textwidth]{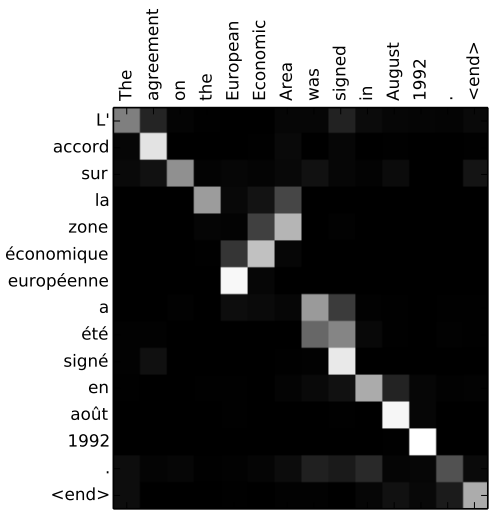}
        \caption{Saliency heatmap \cite{bahdanau:iclr15}}
        \label{fig:saliency-heatmap}
    \end{subfigure}%
    ~~~
    \begin{subfigure}[t]{0.3\textwidth}
        \centering
        \includegraphics[width=\textwidth]{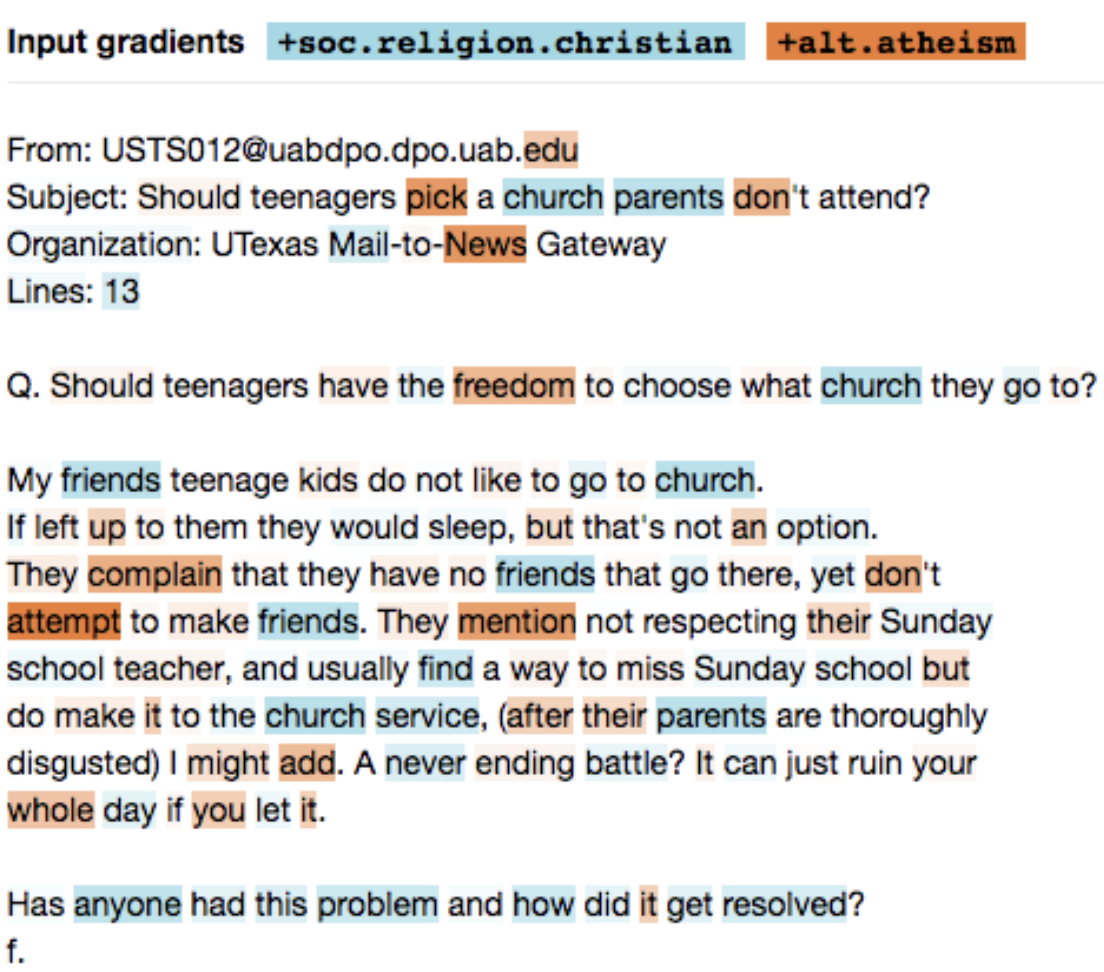}
        \caption{Saliency highlighting \cite{mullenbach-etal-2018-explainable}}
        \label{fig:saliency-highlight}
    \end{subfigure}
    ~~~
    \begin{subfigure}[t]{0.28\textwidth}
        \centering
        \includegraphics[width=\textwidth]{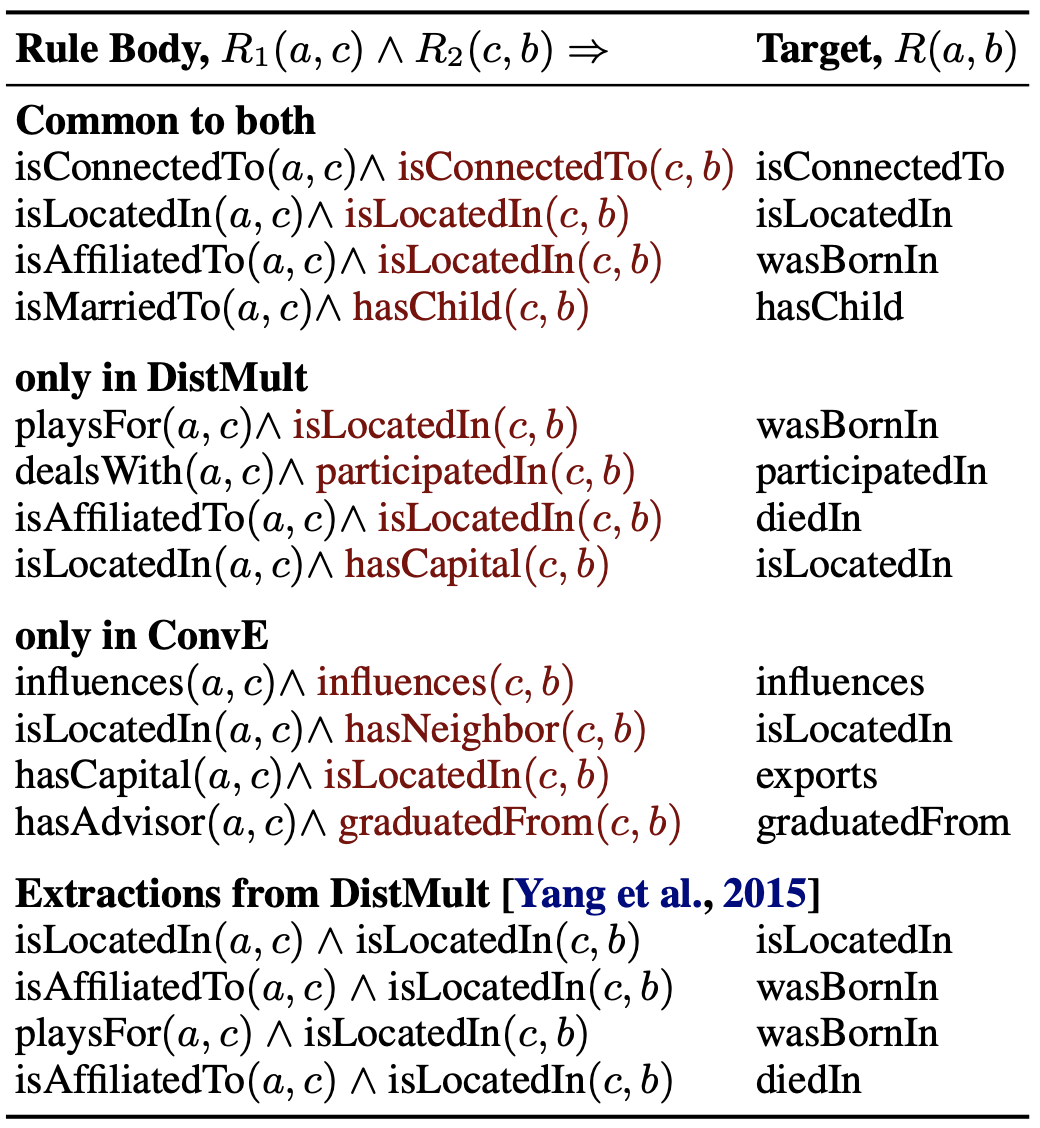}
        \caption{Raw declarative rules \cite{pezeshkpour2019investigating}}
        \label{fig:raw-rule}
    \end{subfigure}
    \\
    \smallskip
    \begin{subfigure}[t]{0.45\textwidth}
        \centering
        \includegraphics[width=\textwidth]{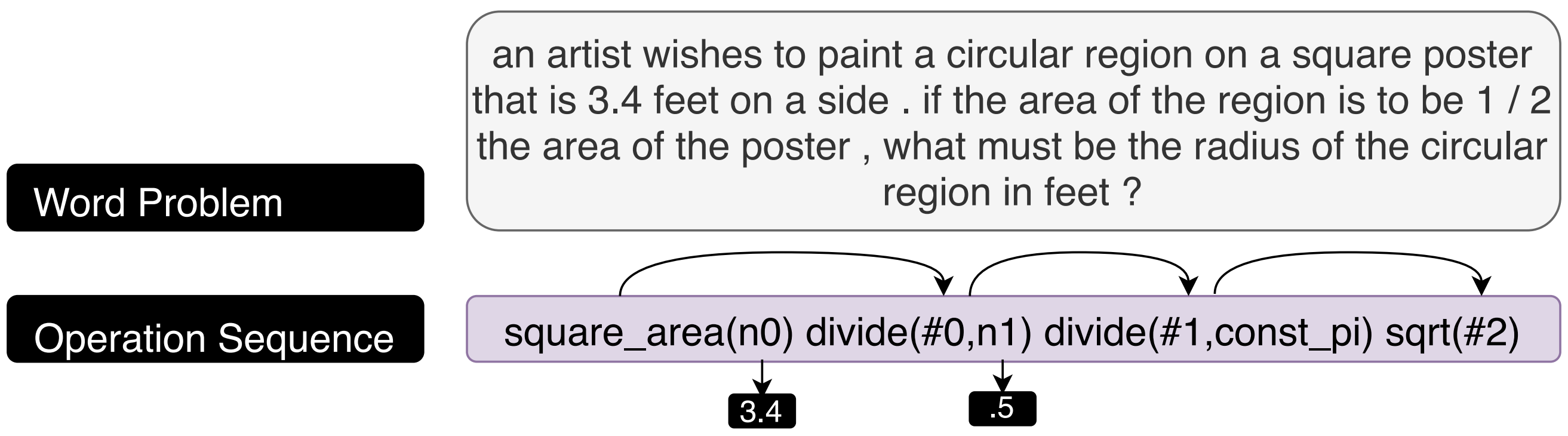}
        \caption{Raw declarative program \cite{amini-etal-2019-mathqa}}
        \label{fig:raw-program}
    \end{subfigure}
    ~~~
    \begin{subfigure}[t]{0.45\textwidth}
        \centering
        \frame{\includegraphics[width=\textwidth]{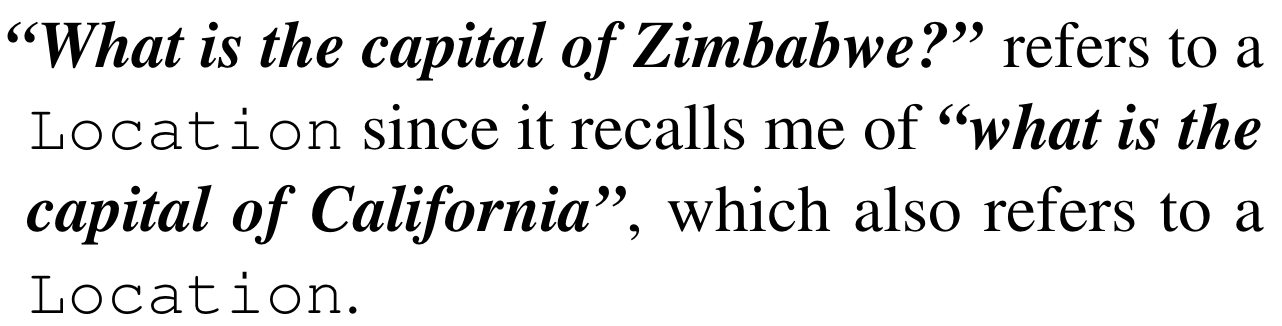}}
        \caption{Raw examples \cite{croce-etal-2019-auditing}}
        \label{fig:raw-examples}
    \end{subfigure}%
    \caption{Examples of different visualization techniques}
    \label{fig:visualization}
\end{figure*}
}

\textit{Saliency.} This has been primarily used to visualize the importance scores of different types of elements in XAI learning systems, such as showing input-output word alignment \cite{bahdanau:iclr15} (Figure \ref{fig:saliency-heatmap}),
highlighting words in input text \cite{mullenbach-etal-2018-explainable} (Figure \ref{fig:saliency-highlight}) or displaying extracted relations \cite{xie-etal-2017-interpretable}.
We observe a strong correspondence between feature importance-based explainability and saliency-based visualizations; namely, 
all papers using feature importance to generate explanations also chose saliency-based visualization techniques. Saliency-based visualizations are popular because they present visually perceptive explanations and can be easily understood by different types of end users. They are therefore frequently seen across different AI domains (e.g., computer vision \cite{simonyan2013deep} and speech \cite{aldeneh2017using}). As shown in Table \ref{tab:aspects-papers}, saliency is the most dominant visualization technique among the papers covered by this survey.

\textit{Raw declarative representations}. As suggested by its name, this visualization technique directly presents the learned declarative representations, such as logic rules, trees, and programs (Figure \ref{fig:raw-rule} and \ref{fig:raw-program}). Such techniques assume that end users can understand specific representations, such as first-order logic rules \cite{pezeshkpour-etal-2019-investigating} and reasoning trees \cite{liang-etal-2016-meaning}, and therefore may implicitly target more advanced users. 

\begin{figure}[ht]
    \centering
    \includegraphics[width=0.42\textwidth]{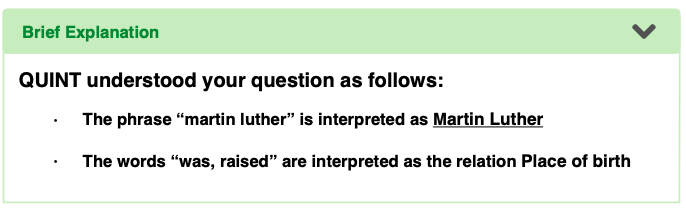}
    \caption{Template-based natural language explanation for a QA system \cite{abujabal2017quint}.}
    \label{fig:natural-language}
    \vspace{-3mm}
\end{figure}

\textit{Natural language explanation}. The explanation is verbalized in human-comprehensible natural language (Figure \ref{fig:natural-language}). The natural language can be generated using sophisticated deep learning models, e.g., by training a language model with human natural language explanations and coupling with a deep generative model \cite{rajani-etal-2019-explain}. It can also be generated by using simple template-based approaches \cite{abujabal2017quint}. In fact, many declarative induction-based techniques can use template-based natural language generation \cite{reiter1997building} to turn rules and programs into human-comprehensible language, and this minor extension can potentially make the explanation more accessible to lay users. 

Table \ref{tab:aspects-papers} references some additional visualization techniques, such as using \textit{raw examples} to present example-driven approaches \citep{jiang-etal-2019-explore, croce-etal-2019-auditing} 
(e.g., Figure \ref{fig:raw-examples}), and dependency parse trees to represent input questions \cite{abujabal2017quint}.


\section{Explanation Quality}
\label{sec:how-well}





Following the goals of XAI, a model's quality should be evaluated not only by its accuracy and performance, but also by how well it provides explanations for its predictions. 
In this section we discuss the state of the field in terms of defining and measuring explanation quality.


\subsection{Evaluation}
\label{sec:evaluation}


Given the young age of the field, unsurprisingly there is little agreement on how explanations should be evaluated. The majority of the works reviewed (32 out of 50) either lack a standardized evaluation or include only an informal evaluation, while a smaller number of papers looked at more formal evaluation approaches, including leveraging ground truth data and human evaluation. We next present the major categories of evaluation techniques we encountered (summarized in Table \ref{tab:eval_techniques}).

\begin{table}[ht]
\small
\centering
\begin{tabular}{ccc} 
\toprule
None or Informal & Comparison to & Human\\
Examination only & Ground Truth & Evaluation\\
\arrayrulecolor{gray} \midrule
32 & 12 & 9 \\
\arrayrulecolor{black} \bottomrule
\end{tabular}
\caption{Common evaluation techniques and number of papers adopting them, out of the 50 papers surveyed (note that some papers adopt more than one technique)}
\label{tab:eval_techniques}
\end{table}

\emph{Informal examination} of explanations. This typically takes the form of high-level discussions of how examples of generated explanations align with human intuition.
This includes cases where the output of a single explainability approach is examined in isolation \cite{xie-etal-2017-interpretable} as well as when explanations are compared to those of other reference approaches \cite{ijcai2017-371} (such as LIME, which is a frequently used baseline).

\emph{Comparison to ground truth.}
Several works compare generated explanations to ground truth data in order to quantify the performance of explainability techniques. 
Employed metrics vary based on task and explainability technique, but commonly encountered metrics include P/R/F1 \cite{carton-etal-2018-extractive}, perplexity, and BLEU \cite{ling-etal-2017-program, rajani2019explain}. 
While having a quantitative way to measure explainability is a promising direction, care should be taken during ground truth acquisition to ensure its quality and account for cases where there may be alternative valid explanations. Approaches employed to address this issue involve having multiple annotators and reporting inter-annotator agreement or mean human performance, as well as evaluating the explanations at different granularities (e.g., token-wise vs phrase-wise) to account for disagreements on the precise value of the ground truth \cite{carton-etal-2018-extractive}.

\emph{Human evaluation.} 
A more direct way to assess the explanation quality is to ask humans to evaluate the effectiveness of the generated explanations. This has the advantage of avoiding the assumption that there is only one good explanation that could serve as ground truth, as well as sidestepping the need to measure similarity of explanations. 
Here as well, it is important to have multiple annotators, report inter-annotator agreement, and correctly deal with subjectivity and variance in the responses.
The approaches found in this survey vary in several dimensions, including the number of humans involved (ranging from 1 \cite{mullenbach-etal-2018-explainable} to 25 \cite{sydorova-etal-2019-interpretable} humans), as well as the high-level task that they were asked to perform (including rating the explanations of a single approach \cite{dong-etal-2019-editnts} and comparing explanations of multiple techniques \cite{sydorova-etal-2019-interpretable}). 

\emph{Other operation-specific techniques.} Given the prevalence of attention layers \citep{bahdanau:iclr15,vaswani:nips17} in NLP, recent work \citep{jain-wallace-2019-attention,serrano-smith-2019-attention,wiegreffe-pinter-2019-attention} has developed specific techniques to evaluate such explanations based on counterfactuals or erasure-based tests \citep{feng-etal-2018-pathologies}. \citeauthor{serrano-smith-2019-attention} repeatedly set to zero the maximal entry produced by the attention layer. If attention weights indeed ``explain" the output prediction, then turning off the dominant weights should result in an altered prediction. Similar experiments have been devised by others \citep{jain-wallace-2019-attention}. In particular, \citeauthor{wiegreffe-pinter-2019-attention} caution against assuming that there exists only one true explanation to suggest accounting for the natural variance of attention layers. On a broader note, causality has thoroughly explored such counterfactual-based notions of explanation \citep{halpern:book16}. 

While the above overview summarizes \emph{how} explainability approaches are commonly evaluated, 
another important aspect is \emph{what} is being evaluated.
Explanations are multi-faceted objects that can be evaluated on multiple aspects, such as \emph{fidelity} (how much they reflect the actual workings of the underlying model), \emph{comprehensibility} (how easy they are to understand by humans), and others. Therefore, understanding the target of the evaluation is important for interpreting the evaluation results.
We refer interested readers to \citep{carvalho_machine_2019} for a comprehensive presentation of aspects of evaluating approaches.


%

Many works do not explicitly state what is being evaluated.
As a notable exception, \cite{lertvittayakumjorn-toni-2019-human} outlines three goals of explanations (reveal model behavior, justify model predictions, and assist humans in investigating uncertain predictions) and proposes human evaluation experiments targeting each of them.

\subsection{Predictive Process Coverage}
\label{sec:part-covered}





An important and often overlooked aspect of explanation quality is
the part of the prediction process (starting with the input and ending with the model output) covered by an explanation. We have observed that many explainability approaches explain only part of this process, leaving it up to the end user to fill in the gaps.

As an example, consider the MathQA task of solving math word problems. As readers may be familiar from past education experience, in math exams, one is often asked to provide a step-by-step explanation of how the answer was derived. Usually, full credit is not given if any of the critical steps used in the derivation are missing. Recent works have studied the explainability of MathQA models, which seek to reproduce this process \cite{amini-etal-2019-mathqa, ling-etal-2017-program}, and have employed different approaches in the type of explanations produced. While \cite{amini-etal-2019-mathqa} explains the predicted answer by showing the sequence of mathematical operations leading to it, this provides only partial coverage, as it does not explain how these operations were derived from the input text. 
On the other hand, the explanations produced by \cite{ling-etal-2017-program} augment the mathematical formulas with text describing \textit{the thought process} behind the derived solution, thus covering a bigger part of the prediction process.

The level of coverage may be an artifact of explainability techniques used: provenance-based approaches tend to provide more coverage, while example-driven approaches, may provide little to no coverage.
Moreover, while our math teacher would argue that providing higher coverage is always beneficial to the student, in reality this may depend on the end use of the explanation. 
For instance, the coverage of explanations of \cite{amini-etal-2019-mathqa} may be potentially sufficient for advanced technical users. Thus, higher coverage, while in general a positive aspect, should always be considered in combination with the target use and audience of the produced explanations.

\section{Insights and Future Directions}
\label{sec:conclusion}

 This survey showcases recent advances of XAI research in NLP, as evidenced by publications in major NLP conferences in the last 7 years.
 We have discussed the main categorization of explanations (Local vs Global, Self-Explaining vs Post-Hoc) as well as the various ways explanations can be arrived at and visualized, together with the common techniques used.
 We have also detailed operations and explainability techniques currently available for generating explanations of model predictions, in the hopes of serving as a resource for developers interested in building explainable NLP models.
 

We hope this survey encourages the research community to work in bridging the current gaps in the field of XAI in NLP.
The first research direction is a need for clearer terminology and understanding of what constitutes explainability and how it connects to the target audience. For example, is a model that displays an induced program that, when executed, yields a prediction, and yet conceals the process of inducing the program, explainable in general? Or is it explainable for some target users but not for others?  
The second is an expansion of the evaluation processes and metrics, especially for human evaluation.
The field of XAI is aimed at adding explainability as a desired feature of models, in addition to the model's predictive quality, and other features such as runtime performance, complexity or memory usage. In general, trade-offs exist between desired characteristics of models, such as more complex models achieving better predictive power at the expense of slower runtime. 
In XAI, some works have claimed that explainability may come at the price of losing predictive quality \cite{Bertsimas2019ThePO}, while other have claimed the opposite \cite{garneau-etal-2018-predicting,liang-etal-2016-meaning}.
Studying such possible trade-offs is an important research area for XAI, but one that cannot advance until standardized metrics are developed for evaluating the quality of explanations.
The third research direction is a call to more critically address the issue of fidelity (or causality), and to ask hard questions about whether a claimed explanation is faithfully explaining the model's prediction.

Finally, it is interesting to note that we found only four papers that fall into the global explanations category. This might seem surprising given that white box models, which have been fundamental in NLP, are explainable in the global sense. We believe this stems from the fact that because white box models are clearly explainable, the focus of the explicit XAI field 
is in explaining black box models, which comprise mostly local explanations. White box models, like rule based models and decision trees, while still in use, are less frequently framed as explainable or interpretable, and are hence not the main thrust of where the field is going. We think that this may be an oversight of the field since white box models can be a great test bed for studying techniques for evaluating explanations.


\nocite{wallace-etal-2018-interpreting,nguyen-AAAI18-16673,ling-etal-2017-program,gupta-schutze-2018-lisa,schwarzenberg-etal-2019-train,harbecke-etal-2018-learning,panchenko-etal-2017-unsupervised,li-etal-2019-cnm,thorne-etal-2019-generating,barbieri-etal-2018-interpretable,jiang-bansal-2019-self,hsu-AAAI18-interpretable,yang-etal-2019-interpretable,bagher-zadeh-etal-2018-multimodal,pappas-popescu-belis-2014-explaining,kang-etal-2017-detecting,tutek-snajder-2018-iterative,garneau-etal-2018-predicting,kiyono-etal-2018-unsupervised,lu-etal-2019-constructing,jiang-etal-2019-explore,moon-etal-2019-opendialkg,pryzant-etal-2018-interpretable,pryzant-etal-2018-deconfounded, bhutani-etal-2018-exploiting, sen-etal-2019-heidl}

\bibliography{anthology,xai-nlp}
\bibliographystyle{acl_natbib}

\appendix

\section{Appendix A - Methodology}
\label{sec:appendix-methodology}

This survey aims to demonstrate the recent advances of XAI research in NLP, rather than to provide an exhaustive list of XAI papers in the NLP community. 
To this end, we identified relevant papers published in major NLP conferences (ACL, NAACL, EMNLP, and COLING) between 2013 and 2019. 
We filtered for titles containing (lemmatized) terms related to XAI, such as ``explainability", ``interpretability", ``transparent", etc.  While this may ignore some related papers, we argue that representative papers are more likely to include such terms in their titles. In particular, we assume that if authors consider explainability to be a major component of their work, they are more likely to use related keywords in the title of their work. Our search criteria yielded a set of 107 papers.

\begin{table}[ht]
\small
\centering
\begin{tabular}{@{}ccc@{}}
\multicolumn{3}{c}{\textbf{Top 3 NLP Topics}}\\ \midrule
{\scriptsize 1} & {\scriptsize 2} & {\scriptsize 3} \\ \midrule
\begin{tabular}[c]{@{}c@{}}\\Question \\Answering \\ (9)\end{tabular} & \begin{tabular}[c]{@{}c@{}}Computational \\ Social Science \&\\ Social Media\\ (6)\end{tabular} & \begin{tabular}[c]{@{}c@{}}Syntax:\\ Tagging, Chunking\\ \& Parsing\\ (6)\end{tabular} \\ \bottomrule
\multicolumn{1}{l}{} & \multicolumn{1}{l}{} & \multicolumn{1}{l}{} \\ 
\multicolumn{3}{c}{\textbf{Top 3 Conferences}} \\ \midrule
{\scriptsize 1} & {\scriptsize 2} & {\scriptsize 3} \\ \midrule
\begin{tabular}[c]{@{}c@{}}EMNLP\\ (21)\end{tabular} & \begin{tabular}[c]{@{}c@{}}ACL\\ (12)\end{tabular} & \begin{tabular}[c]{@{}c@{}}NAACL\\ (9)\end{tabular} \\ \bottomrule
\end{tabular}
\caption{Top NLP topics and conferences (2013-2019) of papers included in this survey}
\label{tab:task_venue}
\end{table}

 During the paper review process we first verified whether each paper truly fell within the scope of the survey; namely, papers with a focus on explainability as a vehicle for understanding how a model arrives at its result. This process excluded 57 papers, leaving us with a total of 50 papers. Table \ref{tab:task_venue} lists the top three broad NLP topics (taken verbatim from the ACL call for papers) covered by these 50 papers, and the top three conferences of the set. 
 
 To ensure a consistent classification, each paper was individually reviewed by at least two reviewers, consulting additional reviewers in the case of disagreement.




\end{document}